# New Worst-Case Upper Bound for X3SAT

Junping Zhou, Minghao Yin

Department of Computer, Northeast Normal University, Changchun, China, 130117

ymh@nenu.edu.cn

**Abstract.** The rigorous theoretical analyses of algorithms for exact 3-satisfiability (X3SAT) have been proposed in the literature. As we know, previous algorithms for solving X3SAT have been analyzed only regarding the number of variables as the parameter. However, the time complexity for solving X3SAT instances depends not only on the number of variables, but also on the number of clauses. Therefore, it is significant to exploit the time complexity from the other point of view, i.e. the number of clauses. In this paper, we present algorithms for solving X3SAT with rigorous complexity analyses using the number of clauses as the parameter. By analyzing the algorithms, we obtain the new worst-case upper bounds $O(1.15855^m)$, where $m$ is the number of clauses.

## 1 Introduction

Exact satisfiability problem, abbreviated XSAT, is a problem of deciding whether there is a truth assignment satisfying exactly one literal in each clause. The exact 3-satisfiability (X3SAT) is the version in which each clause contains at most three literals. The X3SAT problem is an important variant of the well-known NP-complete problem of propositional satisfiability (SAT), which has played a key role in complexity theory as well as in automated planning. In fact, X3SAT is also a NP-complete problem even when restricted to all variables occurring only unnegated [1]. If P ≠ NP, it means that we can't solve the problem in polynomial time. Therefore, Improvements in the exponential time bounds are crucial in determining the size of NP-complete class problem that can be solved. Even a slight improvement from $O(c^k)$ to $O((c-\varepsilon)^k)$ may significantly increase the size of the problem being tractable.

Recently, tremendous efforts have been made on analyzing of algorithms for X3SAT problems. Based on a recursive partitioning of the problem domain and a careful elimination of some branches, Drori and Peleg presented an algorithm running in $O(1.1545^n)$ for X3SAT, where $n$ is the number of the variables [2]. By adapting and improving branching techniques, Porschen et al. proposed an algorithm for solving X3SAT running in $O(1.1487^n)$ [3]. According to exploit a perfect matching reduction and present a more involved deterministic case analysis, Porschen et al. prove a new upper bound for X3SAT ($O(1.1926^n)$) [4]. By providing a new transformation rule, Kulikov [5] simplified the proof of the bound for X3SAT ($O(1.1926^n)$) presented by Porschen et al. [4]. Based on combining various techniques including matching and reduction, Dahllöf et al. addressed an algorithm running in $O(1.1120^n)$ for X3SAT [6]. Further improved algorithms in [7] presented a new upper time bound for the X3SAT ($O(1.1004^n)$), which is the best upper bound so far.

Different from complexity analyses regarding the number of variables as the parameter, Skjernaa presented an algorithm for XSAT with a time bound $O(2^m)$ but using exponential space, where $m$ is the number of clauses of a formula [8]. Bolette addressed an algorithm for XSAT with polynomial space usage and a time bound $O(m!)$ [9]. Similar to the XSAT problem, the time complexity of X3SAT problem is calculated based on the size of the X3SAT instances, which depends not only on the number of variables, but also on the number of clauses. Therefore, it is significant to exploit the time complexity from the other point of view, i.e. the number of clauses. However, so far all algorithms for solving X3SAT have been analyzed based on the number of variables. And to our best knowledge, it is still an open problem that analyzes the X3SAT algorithm with the number of clauses as the parameter.

The aim of this paper is to exploit new upper bounds for X3SAT using the number of clauses as the parameter. We provide algorithm for solving X3SAT. This algorithm employs a new principle, i.e. the connected-clauses principle, to simplify formulae. This allows us to remove one sub-formula and therefore reduce as many clauses as possible in both branches. In addition, by improving the case analyses, we obtain the worst-case upper bound for solving X3SAT is $O(1.15855^m)$, where $m$ is the number of clauses of a formula.

## 2   Problem Definitions

We describe some definitions used in this paper. A variable can take the values *true* or *false*. A literal of a variable is either the unnegated literal $x$, having the same truth value as the variable, or the negated literal $\neg x$, having the opposite truth value as the variable. A clause is a disjunction of literals, referred to as a *k-clause* if the clause is a disjunction on $k$ literals. A *k*-SAT formula $F$ in Conjunction Normal Form (CNF) is a conjunction of clauses, each of which contains at most $k$ literals. A truth assignment for $F$ is a map that assigns each variable a value. When a truth assignment makes the $F$ true, we say the truth assignment is a satisfying assignment. The *exact satisfiability problem* (XSAT) is to find a truth assignment such that exactly one literal is *true* in each clause. The *exact 3-satisfiability problem* (X3SAT) is a version of the XSAT in which each clause contains at most three literals. We define $m$ as the number of clauses in $F$, and $n$ as the number of variables $F$ contains. When a variable occurs once in $F$, it is referred to as singleton. The degree of a variable $v$, represented by $\varphi(v)$, is the number of times it occurs in a formula. The degree of a formula $F$, denoted by $\varphi(F)$, is the maximum degree of variables in $F$. A literal $x$ is an $(i, j)$-literal if $F$ contains exactly $i$ occurrences of $x$ and exactly $j$ occurrences of $\neg x$. And a literal $x$ is monotone if its complementary literal does not appear in $F$. Given a literal $x$, we say

var(x) is the variable that forms the literal and ~ x indicates x or ¬x. We also use $F(\mu \leftarrow \eta)$ to denote the substitution of $\mu$ by $\eta$ in the formula F, where $\mu$ is either a literal or a clause, and $\eta$ is a literal, clause, or *false*. To avoid a tedious enumeration of trivialities, if more than one literal is substituted by *false*, $\mu$ is usually expressed as a set of literals. We use $F / \pi$ to denote the formula obtained by removing $\pi$ from F, where $\pi$ is either a clause or a sub-formula. Given a formula F and a literal x, NumC(F, N(x)) is defined as follow.

$$NumC(F, N(x)) = |\{C | C \in F \wedge var(C) \cap N(x) \neq \varnothing\}| \quad (1)$$

where N(x) is the set of variables that appear in a clause with the literal x, and var(C) is the set of variables occur in the clause C.

After specifying the definitions, we present some basic rules for solving X3SAT problem. Given a formula F, the basic strategy of Davis-Putnam-Logemann-Loveland (DPLL) is to arbitrarily choose a variable v that appears in F. Then,

$$F = (F \wedge v) \vee (F \wedge \neg v) \quad (2)$$

Given a formula F, if F can be partitioned into disjoint sub-formulae where any two sub-formulae have no common variables, then

$$F = F_1 \wedge F_2 \wedge \ldots \wedge F_k \quad (3)$$

Thus, F can be evaluated by deciding the satisfiability of disjoint sub-formulae of F respectively.

### 2.1 Estimating the Running Time

In this subsection, we explain how to compute an upper bound on the running time of a DPLL-style algorithm. At first, we present a notion called branching tree. The branching tree is a hierarchical tree structure with a set of nodes, each of which is labeled with a formula [10]. Suppose there is a node labeled with a formula F, then its sons labeled with $F_1, F_2, \ldots, F_k$ are obtained by branching on one or more variables in the formula F, i.e., assigning values to the variable(s) such that the formula F is reduced to two or more sub-formulae $F_1, F_2, \ldots, F_k$ with fewer variables. Indeed, the construction of a branching tree can be viewed as an execution of a DPLL-style algorithm. Therefore, we use the branching tree to estimate the running time of our algorithm.

In the branching tree, every node has a branching vector. Let us consider a node labeled with $F_0$ and its sons labeled with $F_1, F_2, \ldots, F_k$. The branching vector of the node labeled with $F_0$ is $(r_1, r_2, \ldots, r_k)$, where $r_i = f(F_0) - f(F_i)$ ($f(F_0)$ is the number of clauses of $F_0$). The characteristic polynomial of the branching vector is defined as follows:

$$h(x) = 1 - \sum_{i=1}^{k} x^{-r_i} \quad (4)$$

The positive root of this polynomial is called the branching number, denoted by $\lambda (r_1, r_2, \ldots, r_k)$. And we assume that the branching number of the leaves is 1. We define the maximum branching number of nodes in the branching tree as the branching number of the branching tree, expressed by max $\lambda$ $(r_1, r_2,..., r_k)$. The branching number of a branching tree has an important relationship with the running time (T($m$)) of DPLL-style algorithms. At first, we assume that the running time of DPLL-style algorithms performing on each node is in polynomial time. Then we obtain the following inequality.

$$T(m) \leq (\max \lambda \ (r_1, r_2,..., r_k))^m \times \text{poly}(F)$$
$$= (\max \sum_{i=1}^{k} T(m-r_i))^m \times \text{poly}(F) \quad (5)$$

where $m$ is the number of clauses in the formula $F$, ploy($F$) is the polynomial time executing on the node $F$, and

$$\lambda \ (r_1, r_2,..., r_k) = \sum_{i=1}^{k} T(m-r_i) \quad (6)$$

In addition, if a X3SAT problem recursively solved by the DPLL-style algorithms, the time required doesn't increase, for

$$\sum_{i=1}^{k} T(m_i) \leq T(m) \text{ where } m = \sum_{i=1}^{k} m_i \quad (7)$$

where $m$ is the number of clauses, $m_i$ is the number of clauses in the sub-formula $F_i$ ($1 \leq i \leq k$) of the formula $F$. Note that when analyzing the running time of our algorithms, we ignore the polynomial factor so that we assume that all polynomial time computations take $O(1)$ time in this paper.

## 3   Algorithm for Solving X3SAT

In this section, we present the algorithm X3SAT and prove an upper bound $O(1.15855^m)$, where $m$ is the number of the clauses. Firstly we address some transformation rules used in the algorithm.

### 3.1   Transformation Rules

The transformation rules are applied before branching on one or more variables of the formula $F$. They can reduce formula such that the simplified formula contains a fewer number of clauses. In the following, we present the transformation rules (TR1) - (TR14) which are also used by [7].

**(TR1).** If $F$ contains a variable $x$ such that the number of negated occurrences is larger than the number of unnegated occurrences, then let $F = F(\neg x \leftarrow x)$.

**(TR2).** If $F$ contains a 1-clause $C = x$, then $F = F(x \leftarrow true)$.

**(TR3).** If $F$ contains a 2-clause $C = x \vee y$, then $F = F(x \leftarrow \neg y)$.

**(TR4).** If $F$ contains a clause $C = x \vee x \vee y$, then $F = F(x \leftarrow false)$.

**(TR5).** If $F$ contains a clause $C = x \vee \neg x \vee y$, then $F = F(y \leftarrow false)$.

**(TR6).** If $F$ contains a clause $C = x \vee y \vee z$ where $x$ and $y$ are singletons, then $F = F(x \leftarrow false)$.

**(TR7).** If $F$ contains clauses $C_1 = x \vee y \vee z$, $C_2 = x \vee \neg y \vee z'$, then $F = F(x \leftarrow false)$.

**(TR8).** If $F$ contains clauses $C_1 = x \vee y \vee z$, $C_2 = \neg x \vee \neg y \vee z'$, then $F = F(y \leftarrow \neg x)$.

**(TR9).** If $F$ contains clauses $C_1 = x_1 \vee y_1 \vee y_2$, $C_2 = x_2 \vee y_2 \vee y_3$, $C_3 = x_3 \vee \neg y_3 \vee y_1$, then $F = F(C_3 \leftarrow (\neg x_1 \vee x_2 \vee x_3))$.

**(TR10).** If $F$ contains clauses $C_1 = x_1 \vee \neg y_1 \vee y_2$, $C_2 = x_2 \vee \neg y_2 \vee y_3$, ..., $C_k = x_k \vee \neg y_k \vee y_1$, then $F = F(\{x_1, x_2, ..., x_k\} \leftarrow false)$.

**(TR11).** If $F$ contains clauses $C_1 = x_1 \vee y_1 \vee y_2$, $C_2 = x_2 \vee y_2 \vee y_3$, $C_3 = x_3 \vee \neg y_3 \vee y_1$ where $x_1$ is a singleton, then $F = F / C_1$.

**(TR12).** If $F$ contains clauses $C_1 = x_1 \vee y_1 \vee y_2$, $C_2 = x_2 \vee y_2 \vee y_3$, $C_3 = x_3 \vee y_3 \vee y_1$ where $val(x_3)$ is a singleton, then $F = F(C_3 \leftarrow (\neg x_1 \vee y_3 \vee x_3))$.

**(TR13).** If $F$ contains clauses $C_1 = x_1 \vee y_1 \vee z_1$, $C_2 = x_1 \vee y_1 \vee z_2$, then $F = F(C_2 \leftarrow (\neg z_1 \vee z_2))$.

**(TR14).** If $F$ contains a clause $C = x \vee y \vee z$, where x and y only occur unnegated and in clauses with a singleton in all other clauses, then $F = F(x_2 \leftarrow false)$.

Actually, the above transformation rules are used in the *Reduce* function repeatedly until no transformation rule applies, which can be guaranteed to terminate in polynomial time. The function takes a CNF $F$ as the input and returns a simplified X3SAT formula. In the following, we will show the character of the simplified X3SAT formula. From now on, unless otherwise stated, given a literal $x$, $Y_1 = \{y_1, y_2, ...\}$ is the set of literals that occur in a clause with $x$; $Y_2 = \{y'_1, y'_2, ...\}$ is the set of literals that occur in a clause with $\neg x$ and $Y = Y_1 \cup Y_2$; $Z = \{z_1, z_2, ...\}$ is the set of literals that don't occur in a clause with $x$. We use $y$'s literals indicating the literals occur in Y. For example, if $x$ is a (2, 1) – literal, the clauses the literal $x$ in are showed in Fig. 1.

$$C_1 = x \vee y_1 \vee y_2 \quad C_2 = x \vee y_3 \vee y_4 \quad C_3 = \neg x \vee y'_1 \vee y'_2$$

**Fig. 1.** The clauses that the literal $x$ appears in when $x$ is a (2, 1) – literal

**Theorem 1[7].** A simplified X3SAT formula contains no 1-clauses or 2-clauses, and no two clauses have more than one variable in common; no clause has more than one singleton; all (a, 0)-literals and (a, 1)-literals that are not singletons are in a clause with no singletons.

**Theorem 2.** If a X3SAT formula contains clauses $C_1 = x \vee y_1 \vee y_2$, $C_2 = x \vee y_3 \vee y_4$, and $C_3 = \neg y_1 \vee y_3 \vee z_1$ where $y_1$ is a (1, 1) - literal, then $F = F((C_1 \wedge C_3) \leftarrow (\neg y_4 \vee y_2 \vee z_1))$ and $\varphi(x) = \varphi(x) - 1$.

*Proof.* If a X3SAT formula contains clauses $C_1 = x \vee y_1 \vee y_2$, $C_2 = x \vee y_3 \vee y_4$, and $C_3 = \neg y_1 \vee y_3 \vee z_1$, then $F = F(C_3 \leftarrow (\neg y_4 \vee y_2 \vee z_1))$ by (TR9). Since $y_1$ is a (1, 1) - literal, $\neg y_1$ is removed by (TR9). Thus, $y_1$ is a singleton in $F$. If a X3SAT formula contains clauses $C_1 = x \vee y_1 \vee y_2$, $C_2 = x \vee y_3 \vee y_4$, and $C_3 = \neg y_4 \vee y_2 \vee z_1$, then we can apply (TR11) and obtain $F = F / C_1$. Therefore, $F = F((C_1 \wedge C_3) \leftarrow (\neg y_4 \vee y_2 \vee z_1))$ and $\varphi(x) = \varphi(x) - 1$. □

**Theorem 3.** When X3SAT formula $F$ contains a clause $y'_1 \vee y_3 \vee z_1$ and a (2, 1) – literal $x$, the formula $F$ can be simplified and the literal $x$ becomes a (2, 0) – literal.

*Proof.* Since $x$ is a (2, 1) – literal, $F$ contains clauses $C_2 = x \vee y_3 \vee y_4$ and $C_3 = \neg x \vee y'_1 \vee y'_2$. Then we can apply (TR9) to $(y'_1 \vee y_3 \vee z_1)$, $(x \vee y_3 \vee y_4)$, and $(\neg x \vee y'_1 \vee y'_2)$, which can transform $F$ to contain $(\neg z_1 \vee y_4 \vee y'_2)$ instead of $(\neg x \vee y'_1 \vee y'_2)$. Therefore, $F$ can be simplified and the literal $x$ becomes a (2, 0) – literal. □

**Theorem 4.** When X3SAT formula $F$ contains a (3, 0) – literal $x$, a singleton $y_4$, and a clause $y_1 \vee y_3 \vee z_1$, the formula $F$ can be simplified and the literal $x$ becomes a (2, 0) – literal.

*Proof.* If $x$ is a (3, 0) – literal, the formula $F$ contains clauses $C_1 = x \vee y_1 \vee y_2$ and $C_2 = x \vee y_3 \vee y_4$. Using the (TR12) on $(x \vee y_1 \vee y_2)$, $(x \vee y_3 \vee y_4)$, and $(y_1 \vee y_3 \vee z_1)$ where $y_4$ is a singleton, we can replace $(x \vee y_3 \vee y_4)$ by $(\neg y_2 \vee y_3 \vee y_4)$. Therefore, the formula $F$ can be simplified and the literal $x$ becomes a (2, 0) – literal. □

### 3.2 Helper Principle

In this subsection, we concentrate on introducing the connected-clauses principle. Before presenting the details, we specify some notions used in this part. Given a simplified X3SAT formula $F$ in CNF, $F$ can be expressed as an undirected graph called connection graph. In the connection graph, the vertexes are the clauses of $F$ and the edges between two vertexes if the corresponding clauses contain the same literal. We say that the clause $C$ is connected with $C'$ if there is an edge connecting the corresponding vertexes in the connection graph. We call such two clauses the connected clauses. The character of connected clauses is showed in the following theorem.

**Theorem 5.** For any two connected clauses $C_1$ and $C_2$, there is only one edge connecting the corresponding vertexes in the connection graph.

*Proof.* In order to prove there is only one edge connecting the corresponding vertexes in the connection graph, we need to prove that $C_1$ and $C_2$ have only one common literal. By (TR2) and (TR3) we know that each clause has exactly three literal in a simplified X3SAT formula. If two clauses have common variables, the common variables must form the same literals based on (TR7) and (TR8). According to (TR13), there is at most only one common literal in any two clauses. Therefore, for any two connected clauses, there is only one edge connecting the corresponding vertexes in the connection graph. □

Let us start to propose the connected-clauses principle. Suppose a connection graph $G$ can be partitioned into two components $G_1$ and $G_2$ where there is only one edge $l$ connecting a vertex in $G_1$ to a vertex in $G_2$, i.e. the formula $F$ corresponding to $G$ is partitioned into two sub-formulae $F_1$ and $F_2$ corresponding to the two components with only one common literal $l$. Then, we can determine the satisfiability of the X3SAT formula $F$ as follows.

$$F = ((F_1 \wedge l) \wedge (F_2 \wedge l)) \ \vee \ ((F_1 \wedge \neg l) \wedge (F_2 \wedge \neg l)) \tag{8}$$

The aim of this principle is to partition the formula $F$ into two sub-formulae. When $F_1$ contains a small number of clauses, it can be solved by exhaustive search in polynomial time. This allows us to remove $F_1$ from $F$ and therefore reduce as many clauses as possible in both branches. The following theorem states that the principle in sound.

**Theorem 6.** The connected-clauses principle is sound.

*Proof.* To prove that the connected-clauses principle is sound, we just to prove after applying the connected-clauses principle do not change the satisfiability of the original formula. Suppose a connection graph $G$ can be partitioned into two components $G_1$ and $G_2$ where there is only one edge $l$ connecting a vertex in $G_1$ to a vertex in $G_2$, i.e. the formula $F$ corresponding to $G$ is partitioned into two sub-formulae $F_1$ and $F_2$ corresponding to the two components with only one common literal $l$. Then after applying the connected-clauses principle to the formula $F$, the formula $F$ can be partitioned into two formulae $F_1$ and $F_2$.

Suppose $F$ is satisfiable. Consider a satisfying assignment $I$ for $F$. It is obvious that in the satisfying assignment the literal $l$ either *true* or *false*. We assume that the literal $l$ is fixed *true*. Then the satisfying assignment for $F$ consists of a satisfying assignment for $F_1 \wedge l$ and a satisfying assignment for $F_2 \wedge l$. The similar situation is encountered when $l$ is fixed *false*.

On the contrary, every satisfying assignment for $F_1 \wedge l$ (resp. $F_1 \wedge \neg l$) can combine with every satisfying assignment for $F_2 \wedge l$ (resp. $F_1 \wedge \neg l$), both of which have an assignment *true* (resp. *false*) for $l$. The combining satisfying assignments are indeed the satisfying assignments for $F$ which has an assignment *true* (*false*) for $l$.

Therefore, the connected-clauses principle is sound. □

### 3.3 Algorithm X3SAT for Solving Exact 3SAT

The algorithm X3SAT for exact 3SAT is based on the DPLL algorithm. The basic idea of the algorithm is to choose a variable and recursively determine whether the formula is satisfiable or not when variable is *true* or *false*. Before presenting the algorithm X3SAT, we address a function $\Omega(F, l)$ in Fig. 2, which recursively executes the propagation. The function takes a formula $F$ and a literal $x$ being assigned *true* as input. The detailed process of the function is presented as follows. (1) Remove all clauses containing literal $x$ from $F$; (2) delete all literals occurring with $x$ from the other clauses; (3) delete all occurrences of the negation of literal $x$ from $F$; (4) perform the process as far as possible.

---
Function $\Omega(F, x)$
1. If there exists a clause $x \vee y_1 \vee y_2$ in $F$,
   then remove the clause $x \vee y_1 \vee y_2$ and the literals $y_1, y_2$ from $F$.
2. If there exists a clause $\neg x \vee y'_1 \vee y'_2$ in $F$, remove $\neg x$ from $\neg x \vee y'_1 \vee y'_2$.
3. For $1 \leq i \leq 2$ do $\Omega(F, \neg y_i)$.
4. Return $F = Reduce(F)$.
---

**Fig. 2.** The function $\Omega$

Now let us start to describe the framework of our algorithm X3SAT in Fig. 3. The algorithm employs a new principle, i.e. the connected-clauses principle, to simplify formulae. It takes a simplified X3SAT formula $F$ as the input. Note that in the algorithm ESX3SAT($F$) is a function that solves the X3SAT by exhaustive search. As we all know, if a X3SAT instance is solved by exhaustive search, it will spend a lot of time. However, when the number of clauses that the formula $F$ contains is so few, it may run in polynomial time. Therefore, we use the function ESX3SAT($F$) only when the number of clauses isn't above 5, which can guarantee the exhaustive search runs in polynomial time. Prefect_Matching($F$) is also a function that reduces the X3SAT instance to a perfect matching problem when $\varphi(F) \leq 2$, and this can be solved in polynomial time [11]. In Theorem 7, we analyze the algorithm X3SAT using the measure described above.

---

Algorithm X3SAT($F$)
Case 1: $F$ has an empty clause. return *unsatisfiable*.
Case 2: $F$ is empty. return *satisfiable*.
Case 3: $m < 6$. return ESX3SAT ($F$).
Case 4: $F$ consists of disjoint sub-formulae $F_1, F_2, \ldots, F_k$.
  return X3SAT ($F_1$) $\wedge$ X3SAT ($F_2$) $\wedge \ldots \wedge$ X3SAT ($F_k$).
Case 5: $\varphi(F) \geq 4$. Pick a maximum degree variable $x$.
return X3SAT ($\Omega(F, x)$) $\vee$ X3SAT ($\Omega(F, \neg x)$).
Case 6: $\varphi(F)=3$ and there is a (2, 1)-literal $x$ such that $C_1 = x \vee y_1 \vee y_2$, $C_2 = x \vee y_3 \vee y_4$, and $C_3 = \neg x \vee y'_1 \vee y'_2$.
1. If two clauses $C_4$ and $C_5$ connect with $C_1 \sim C_3$.
(1) $C_4$ connects with $C_1$ and $C_2$, $C_5$ connects with $C_3$, i.e., $C_4 = \sim y_1 \vee \sim y_3 \vee z_1$, and $C_5 = \sim y'_1 \vee z_2 \vee z_3$, where $z_1$ is a singleton.
  return X3SAT( $\Omega(F_1, y'_1) \wedge \Omega(F_2, y'_1)$) $\vee$ X3SAT( $\Omega(F_1, \neg y'_1) \wedge \Omega(F_2, \neg y'_1)$), where $F_1 = C_1 \wedge C_2 \wedge C_3 \wedge C_4$, $F_2 = F/F_1$.
(2) $C_4$ connects with $C_1$ and $C_2$, $C_5$ connects with $C_3$, i.e., $C_4 = \sim y_1 \vee \sim y_3 \vee z_1$, and $C_5 = \sim y'_1 \vee z_2 \vee z_3$, where $z_1$ is not a singleton.
  return X3SAT ($\Omega(F, x)$) $\vee$ X3SAT ($\Omega(F, \neg x)$).
(3) $C_4$ connects with $C_1$; $C_5$ connects with $C_2$ and $C_3$, i.e., $C_4 = \sim y_1 \vee z_1 \vee z_2$ and $C_5 = \sim y'_1 \vee \sim y_3 \vee z_3$, where $z_3$ is a singleton.
  return X3SAT($\Omega(F_1, y_1) \wedge \Omega(F_2, y_1)$) $\vee$ X3SAT($\Omega(F_1, \neg y_1) \wedge \Omega(F_2, \neg y_1)$), where $F_1 = C_1 \wedge C_2 \wedge C_3 \wedge C_5$, $F_2 = F/F_1$.
(4) $C_4$ connects with $C_1$; $C_5$ connects with $C_2$ and $C_3$, i.e., $C_4 = \sim y_1 \vee z_1 \vee z_2$ and $C_5 = \sim y'_1 \vee \sim y_3 \vee z_3$, where $z_3$ is not a singleton.
  return X3SAT ($\Omega(F, x)$) $\vee$ X3SAT ($\Omega(F, \neg x)$).
(5) otherwise, return X3SAT ($\Omega(F, x)$) $\vee$ X3SAT ($\Omega(F, \neg x)$).
2. If three or more clauses connect with $C_1 \sim C_3$.
return X3SAT ($\Omega(F, x)$) $\vee$ X3SAT ($\Omega(F, \neg x)$).
Case 7: $\varphi(F)=3$ and there is a (3, 0)-literal $x$.
return X3SAT ($\Omega(F, x)$) $\vee$ X3SAT ($\Omega(F, \neg x)$).
Case 8: $\varphi(F) \leq 2$, return Prefect_Matching($F$).

---

**Fig. 3.** The algorithm for solving X3SAT

**Theorem 7.** Algorithm X3SAT runs in $O(1.15855^m)$ time, where $m$ is the number of the clauses.

*Proof.* Let us analyze the algorithm case by case.
  Case 1, 2 and 3 can solve the instances completely and run in $O(1)$.
  Case 4 doesn't increase the time needed.

Case 5: When $x=true$, every clause containing $x$ is removed and $\neg x$ is removed from clauses. More over, every clause containing $\neg x$ shrinks to 2-clause which can be removed by (TR3). Therefore, the current formula contains at least four clauses less than $F$ and the same situation is encountered when $x=false$. In addition, when $x$ is fixed a value, the clauses containing the literals in Y can be also removed. Now we let $R= NumC(F, N(x))$ and $R'= NumC(F, N(\neg x))$. Then we have $T(m) = T(m-4-R)+ T(m-4-R')$. By Theorem 1, we know that at least four literals in Y occur in other clauses. So we obtain $R+ R' \geq 2$. Therefore, the worst case is when $T(m)=T(m-6)+T(m-4)$ with solution $O(1.15096^m)$.

Case 6.1.1: When $z_1$ is a singleton, the formula $F$ can be partitioned into two formulae $F_1= C_1 \wedge C_2 \wedge C_3 \wedge C_4$ and $F_2=F/F_1$ with only one common literal $y'_1$. By the connected-clauses principle, we branch on the common literal $y'_1$. We know that when the number of clauses that a formula contains is less than 6, the formula can be solved by exhaustive search. This means that the formula $F_1$ can be solved in polynomial time. And when $y'_1$ is fixed a value, at least one clause containing $y'_1$ is removed from the formula $F_2$. So the current formulae contain at least five clauses less than $F$ in both of the branches. Therefore, we have $T(m)=T(m-5)+ T(m-5)$ with solution $O(1.14870^m)$.

Case 6.1.2: In this case, the $y$'s literals in $C_4$ must be unnegated based on Theorem 2. Thus, when $x=true$, every clause containing $x$ or $z_1$ is removed and every clause containing $\neg x$ or $\neg z_1$ is also removed by (TR3). Since var($z_1$) occurs at least twice and var($x$) occurs three times in $F$, the current formula contains at least five clauses less than $F$. When $x=false$, every clause containing var($x$) or var($y'_1$) can be removed, which make $y_1$ and $y_3$ become singletons. So clause $C_4$ can be removed by (TR6). Therefore, the worst case is when $T(m)=T(m-5)+T(m-5)$ with solution $O(1.14870^m)$.

Case 6.1.3: This case is similar to the case 6.1.1. So the current formula contains at least five clauses less than $F$ in both of the branches. Therefore, we have $T(m)=T(m-5)+ T(m-5)$ with solution $O(1.14870^m)$.

Case 6.1.4: In this case, at least one of the $y$'s literals in $C_5$ must be negated based on Theorem 3. If we give $true$ to $x$, at least four clauses containing var($x$) or var($y_1$), are removed. And simultaneously other clauses containing var($z_3$) are removed. As we know, $z_3$ is not a singleton and this means that var($z_1$) occurs at least twice. So the current formula contains at least six clauses less than $F$. When $x=false$, at least four clauses containing var($x$) or var($y'_1$) are removed. Therefore, the worst case is when $T(m)=T(m-6)+T(m-4)$ with solution $O(1.15096^m)$.

Case 6.1.5: Due to previous cases, we know that $C_4$ and $C_5$ both contain at least two $y$'s literals. When $x=true$, every clause containing $x$ or $y_i$ ($1 \leq i \leq 4$) is removed. When $x=false$, every clause containing $\neg x$ or $y'_j$ ($1 \leq j \leq 2$) is removed. In addition, by Theorem 3, at least one of the $y$'s literals in the clause $C_4$ or $C_5$ with $y'_j$ ($1 \leq j \leq 2$) must be negated and therefore at least two clauses containing literals in $Z$ can be also removed. Thus, it follows that $T(m)=T(m-6)+T(m-4)$ with solution $O(1.15096^m)$.

Case 6.2: Let us assume that $R= NumC(F, N(x))$ and $R'= NumC(F, N(\neg x))$. Since $\varphi(x)=3$, the current formula contains at least three clauses less than $F$ when $x$ is fixed a value. Furthermore, when $x=true$, $y_i=false$ ($1 \leq i \leq 4$) and the clauses containing $y_i$ ($1 \leq i \leq 4$) are removed by (TR3). The time needed in this case is thus bounded by $T(m) = T(m-3-R)+ T(m-3-R')$ since exactly the similar situation arises when $x$ is given the value $false$. It is easy to see that $R \geq 1$ and $R' \geq 1$ for there are three or more clauses connect with $C_1 \sim C_3$. Moreover, at least four literals in Y occur in the three or more clauses by Theorem 1. Consequently, $R + R' \geq 4$ and the worst case occurs when $R = 3$, $R' = 1$. Therefore, The time needed in this case is bounded by $T(m)=T(m-6)+T(m-4)$ and $T(m) \in O(1.15096^m)$.

Case 7: If $x$ is a (3, 0)-literal, at least four variables in $Y_1$ must occur in other clauses by Theorem 1. And if $F$ contains an unnegated and negated variable, it must be (1, 1)-literal, otherwise, the (2, 1)-literal case is met. Therefore, there are at least two clauses connected with $C_1 \sim C_3$. In the following, we analyze the complexity from three cases. (1) Two clauses $C_4$ and $C_5$ connect with $C_1 \sim C_3$. If $F$ contains a clause with three variables in $Y_1$, then $x$ must be given the value *false*. Otherwise, $F$ can be simplified by Theorem 2 and 4. Therefore, the case (1) can be solved in $O(1)$. (2) Three clauses $C_4$, $C_5$, and $C_6$ connect with $C_1 \sim C_3$. Similarly, when $F$ contains a clause with three variables in $Y_1$, the formula $F$ can be solved in $O(1)$. When $C_i$ ($4 \leq i \leq 6$) contains two $y$'s literals, the literals must be unnegated according to Theorem 2 and 4. So when each clause $C_i$ ($4 \leq i \leq 6$) contains two $y$'s literals, we branch on $x$. If $x=true$, three clauses containing $x$ are removed and three clauses containing $y$'s variables are also removed. If $x=false$, we substitute $\neg y_2$ for $y_1$; substitute $\neg y_4$ for $y_3$; and substitute $\neg y_6$ for $y_5$. Consequently, we obtain a formula $F$ contains $(\neg y_2 \lor \neg y_4 \lor z_1)$, $(y_2 \lor \neg y_6 \lor z_2)$, and $(y_6 \lor y_4 \lor z_3)$. It is easy to see that the three clauses can be removed by (TR10). And when there is a clause containing only one $y$'s literal, the clause can be removed by (TR6 and TR4) when $x=false$. Therefore, The time needed in this case is bounded by $T(m)=T(m-6)+T(m-4)$ and $T(m) \in O(1.15096^m)$. (3) Four or more clauses connect with $C_1 \sim C_3$. In this case, we branch on $x$. When $x$ is fix a value, the clauses containing $x$ are removed. And the clauses containing $y$'s variables are removed when $x=true$. Therefore, this case is bounded by $T(m)=T(m-7)+T(m-3)$ and takes $O(1.15855^m)$ time.

Case 8: This case can solve the problems completely and run in $O(1)$.

In total, algorithm X3SAT runs in $O(1.15855^m)$ time, where $m$ is the number of the clauses. □

## 5  Conclusion

This paper addresses the worst-case upper bound for the X3SAT problem with the number of clauses as the parameter. The algorithm presented is a DPLL-style algorithm. In order to improve the algorithms, we put forward a new connected-clauses principle to simplify the formulae. After a skillful analysis of these algorithms, we obtain the worst-case upper bound $O(1.15855^m)$ for X3SAT.

## Acknowledgments

This research is fully supported by the National Natural Science Foundation of China under Grant No.60803102, and also funded by NSFC Major Research Program 60496321: Basic Theory and Core Techniques of Non Canonical Knowledge.